\documentclass[letterpaper, 10 pt, conference]{ieeeconf}  

\IEEEoverridecommandlockouts                              
\usepackage{graphicx} 
\usepackage{float}
\usepackage{subfigure}
\usepackage[noadjust]{cite}
\usepackage{amsmath}
\usepackage{amsmath} 
\usepackage{color}
\usepackage{url}
\usepackage{algorithm}
\usepackage[noend]{algpseudocode}

\title{\LARGE \bf
UAV-AdNet: Unsupervised Anomaly Detection using Deep Neural Networks for Aerial Surveillance
}

\author{Ilker Bozcan and Erdal Kayacan
\thanks{I. Bozcan and E. Kayacan are with the Artificial Intelligence in Robotics Laboratory (Air Lab), Department of Engineering, Aarhus University,
        8000 Aarhus C, Denmark
        {\tt\small \{ilker, erdal\} at eng.au.dk}}%
}

\begin{document}

\maketitle
\thispagestyle{empty}
\pagestyle{empty}

\begin{abstract}

Anomaly detection is a key goal of autonomous surveillance systems that should be able to alert unusual observations. In this paper, we propose a holistic anomaly detection system using deep neural networks for surveillance of critical infrastructures (e.g., airports, harbors, warehouses) using an unmanned aerial vehicle (UAV). First, we present a heuristic method for the explicit representation of spatial layouts of objects in bird-view images. Then, we propose a deep neural network architecture for unsupervised anomaly detection (UAV-AdNet), which is trained on environment representations and GPS labels of bird-view images jointly. Unlike studies in the literature, we combine GPS and image data to predict abnormal observations. We evaluate our model against several baselines on our aerial surveillance dataset and show that it performs better in scene reconstruction and several anomaly detection tasks. The codes, trained models, dataset, and video will be available at \url{https://bozcani.github.io/uavadnet}.


\end{abstract}

\section{INTRODUCTION}

Anomaly detection (i.e., finding unusual patterns that are different from the majority of observations \cite{zimek2017outlier}) is a crucial process for autonomous surveillance systems. These systems should be able to alert humans in case of the existence of suspicious observations during monitoring wide areas. Furthermore, for a higher level of automation, they should be able to learn anomalies without human supervision. 

By the advance in on-board sensor technology and computing power, unmanned aerial vehicles (UAVs) with mounted cameras are extensively used for different visual surveillance applications such as detection of suspicious social events \cite{mehran2009abnormal}, violent human actions \cite{gao2016violence} and vehicles \cite{bozcan2020air}. However, surveillance with UAVs requires human supervision even for assistance in the anomaly detection process. To increase the autonomy level of UAVs for anomaly detection task, (i) an environment representation that allows UAVs to be aware of the existence and spatial locations of objects in a given environment, and (ii) unsupervised learning of anomalies in environment representations are required.

\begin{figure} [!hbt]
\centerline{
\includegraphics[width=0.49\textwidth]{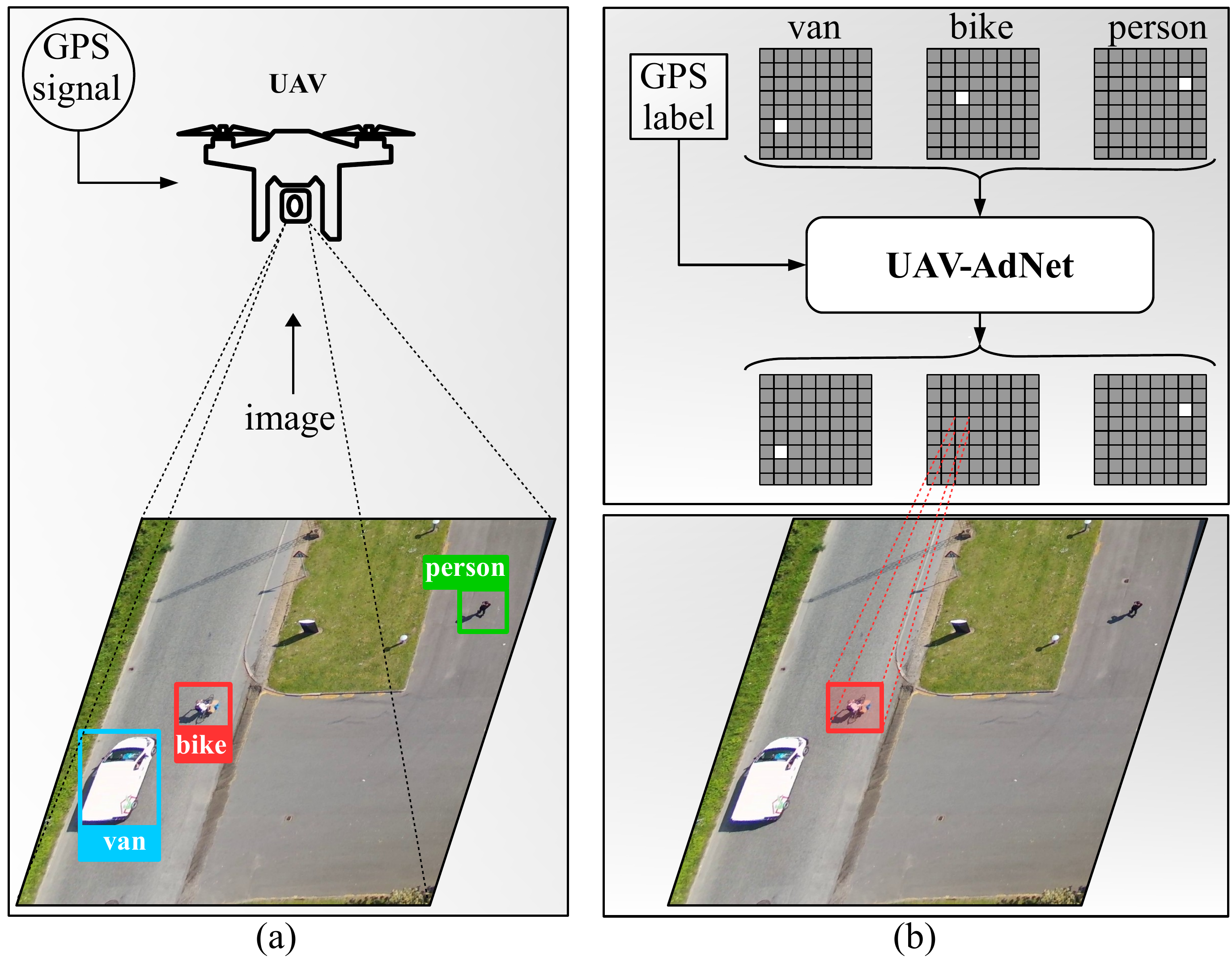}
}
\caption{This figure shows the schematic representation of the proposed anomaly detection framework. Having trained the model, for inference, UAV records a bird-view image and GPS data over the inspection area. After object detection (Fig. 1.a), grid representations are created for an environment. Then, UAV-AdNet recovers the representations for given grids and GPS labels. The difference between the original and reconstructed samples indicates anomalies (e.g., the bike is an anomaly in Fig. 1.b) The UAV-AdNet runs 30 ms on average for inference on the NVIDIA Jetson TX2 \cite{franklin2017nvidia}, which is suitable for real-world applications.\label{fig:intro}}
\end{figure}

In this paper, firstly, we propose a grid-wise environment representation for bird-view images captured by a UAV.  This representation establishes the layout of a scene that shows objects and their spatial locations in the image (See Fig. \ref{fig:intro}). The perspective in aerial images is foreshortened that makes the object appear short and squat, unlike linear perspective in side-view images (Fig. \ref{fig:intro}.a). Therefore, objects in complete bird-view aerial images can be represented in 2D space by discarding the depth, while keeping the altitude of UAV constant. Secondly, we propose an end-to-end trainable neural network architecture for anomaly detection (UAV-AdNet). Similar to variational autoencoders \cite{kingma2013auto}, latent vector-space is constrained for being continuous to learn data distribution. However, we sacrifice generativity for the sake of the precision of predictions by feeding input data in the last layer of the network. 

A presence of an object can be considered anomaly according to the environment that it is presented. For example, a car might be considered normal if it is parked on a car-park, yet it might be anomaly if it appears at the back side of a building. To capture location-dependencies of anomalies, we also feed the network with global positioning system (GPS) data. Therefore, the network learns the data distribution conditioned on flight location (Fig. \ref{fig:intro}.b). We compare the proposed method with several baselines, and show that our method performs the best for anomaly detection tasks.




\subsection{Related Work}
\subsubsection{Environment representation for robots} Environment representation (scene modeling) is crucial for robots that need reasoning and decision making about their environment. Although many scene models have been proposed in robotics using Markov or conditional random fields \cite{CelikkanatConceptWeb2014, anand2013contextually}, Bayesian Networks \cite{li2017context}, Dirichlet and Beta processes \cite{joho2013nonparametric}, predicate logic \cite{mastrogiovanni2011robots}, Scene Graphs \cite{blumenthal2014towards} ontology \cite{saxena2014robobrain,tenorth2009knowrob} and Boltzmann Machines \cite{bozcan2019cosmo, bozcan2018missing}, they have been implemented for indoor tasks using unmanned ground vehicles, humanoid robots or service robots. The high payload capacity of these robots allows scene models to have a variety of input data modalities such as images, point clouds, haptic. Moreover, most of these works have focused on indoor tasks that robots perceive their environment in the frontal (side) view visually. Lack of payload capacity and bird view perception of the environment require unique environment representation for drones.

\subsubsection{Autonomous anomaly detection} Anomaly detection has been studied wildly in variety of domains such as computer vision \cite{gao2016violence, mohammadi2016angry, hasan2016learning, xu2015learning}, computer audition \cite{koizumi2018unsupervised,kawaguchi2017can,foggia2015audio}, finance \cite{wang2017identity, paula2016deep}. Among these works, we highlight three of them which are closely related to our methodology. Xu et al. \cite{xu2015learning} use stacked denoising auto-encoders to learn appearance and motion features and Support Vector Machines (SVM) to predict the anomaly scores for input images. Hasan et al. \cite{hasan2016learning} propose a fully connected autoencoder that has a higher reconstruction error for input images, which include abnormal observations. Koizumi et al. \cite{koizumi2018unsupervised} also use the reconstruction error of autoencoders as a clue for anomalies for audial surveillance. However, in these studies, the reconstruction error is considered an anomaly score. This prevents explicit representation of anomalies (i.e., the spatial layout of anomalies in a sample data). Compared to others, our method can represent each anomaly observation explicitly for a given input (i.e., it can predict both type and location of anomaly objects).

\subsection{Contributions}

The main contributions of our work are the following:
\begin{itemize}
\item \textbf{Environment representation for aerial images:} We propose grid-wise environment modeling, which represents the presence and locations of objects in aerial images. To the best of our knowledge, this is the first study that introduces environment representation for aerial images.


\item \textbf{UAV-AdNet - Deep neural network (DNN) for anomaly detection with UAVs:} We propose a novel DNN architecture, which is dedicated to anomaly detection in environment representations. Our model has crop-and-copy connection, which propagates input data to the late hidden layer directly instead of activation maps, unlike other well-known DNN architectures. We observe that convolving activation maps and inputs together recover spatial information in original input, whereas anomaly parts in original data are removed. Moreover, we train the network with environment representations and GPS labels jointly, and observe that the network learns which objects are anomalies conditioned on GPS labels.

\end{itemize}



We evaluate UAV-AdNet on different surveillance scenarios which UAVs can operate: (i) Determining anomalies which violates private rules (e.g. entering forbidden area of private buildings), (ii) anomalies which violate public rules (e.g. parking car on bike road), (iii) rare observations which may arise suspicions (e.g. truck in a car park). We compare our model (UAV-AdNet) with several baselines including Variational Autoencoders (VAE) \cite{kingma2013auto} and conditional variational autoencoder (CVAE) \cite{sohn2015learning} which are heavily used for unsupervised anomaly detection tasks. During the benchmark evaluation, we exclude classical machine learning approaches for the sake of space and only focus on state-of-the-art deep learning models which are compatible with our proposed method.

The remainder of the paper is organized as follows. Section II introduces the proposed framework for environment representation of aerial images and anomaly detection network (UAV-AdNet). Section III presents the evaluation of the UAV-AdNet on several anomaly detection scenarios and its reconstruction capacity. Section IV discusses the weakness and future of this work. Section V summarizes the work with a conclusion.

\section{PROPOSED METHOD}

In this section, we present environment representation for aerial images that are captured by UAVs and DNN-based anomaly detection technique (UAV-AdNet). As shown in Fig \ref{fig:architecture}, an anomaly detection framework starts by assuming that off-the-shelf object detectors give object annotations for raw input images. Then, grid representations that show the existence and locations of objects are computed using object annotations. Grid representations and corresponding GPS labels which include latitude and longitude coordinates of images are fed to the model as input. The model produces reconstructed data for given input data and GPS label. During the inference phase, the difference between original input data and reconstructed data indicates objects which are an anomaly in the environment. The UAV-AdNet runs 30 ms in average for inference on the NVIDIA Jetson TX2 card, which is suitable for real-world applications.



\subsection{Object Detector}
Initially, we use an object detector to find objects in images. Since object detection is out of the main focus in our work, we use an off-the-shelf object detector. In our experiments, we choose MobileNetV2-SSDLite \cite{sandler2018mobilenetv2} since it is a light-weight object detector that is applicable for real-time on-board applications. 

We implement the same network architecture except for the output layer in the original work \cite{sandler2018mobilenetv2} and train the network using the VisDrone dataset \cite{zhu2018visdrone}, consisting of aerial images and bounding box annotations for different object categories. The annotations for seven object categories (car, pedestrian, bus, van, truck, bicycle, motorbike) are directly inherited from the VisDrone dataset. ``Car'' instances are manually labeled with the ``Trailer'' label if a trailer exists with a car, before training.

The output layer of the network is constructed according to 8 numbers of object categories. We use the standard RMSProp optimizer \cite{tieleman2012lecture}  with momentum set to 0.9. The initial learning rate and batch size are set to 0.05 and 32, respectively. Early-stopping is applied to finish the training.

\subsection{Grid Representation of Environment}
\begin{figure*}[hbt]
\minipage{0.24\textwidth}
  \includegraphics[width=\linewidth]{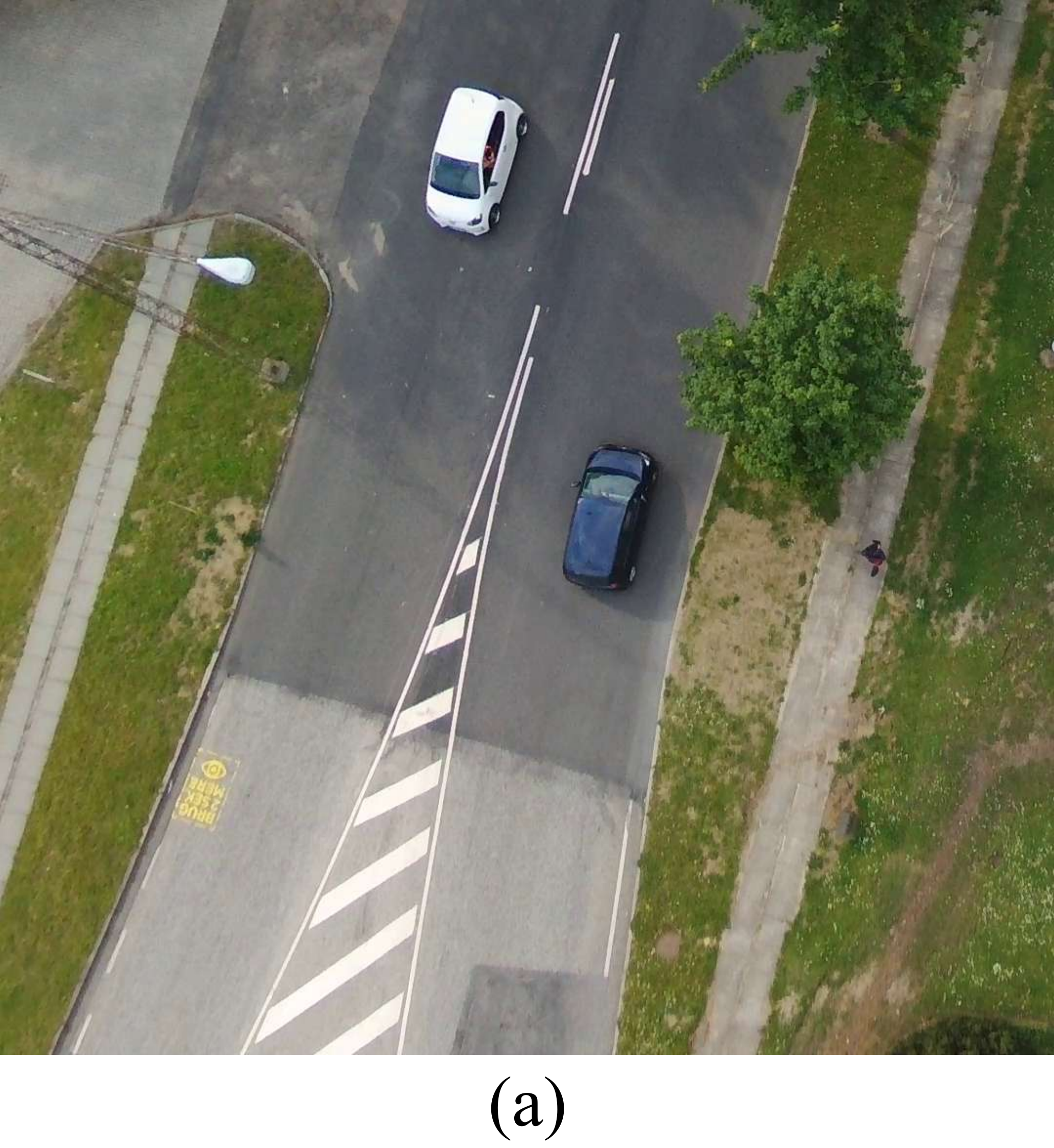}\label{fig:scene_model_a}
\endminipage\hfill
\minipage{0.24\textwidth}
  \includegraphics[width=\linewidth]{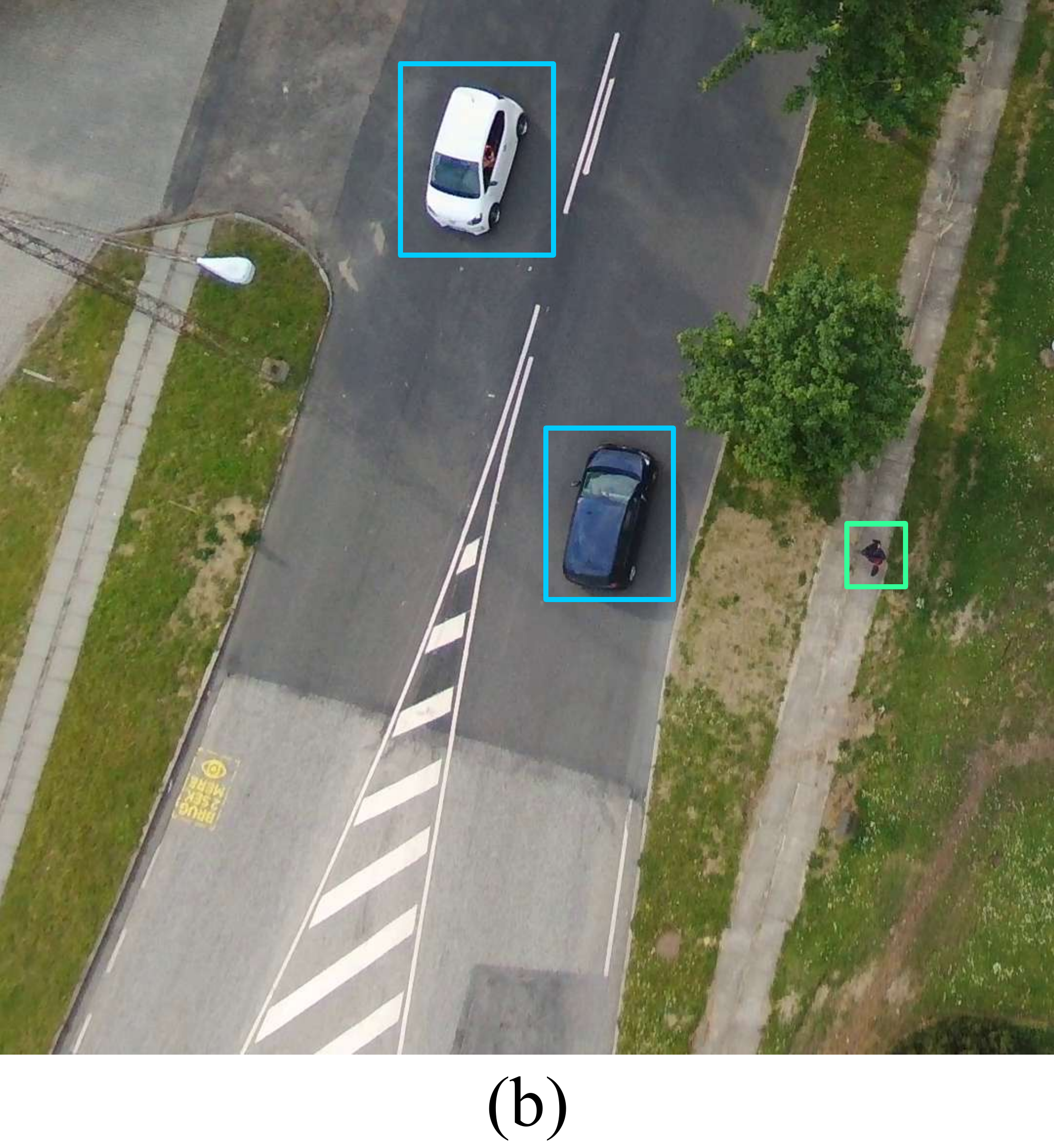}\label{fig:scene_model_b}
\endminipage\hfill
\minipage{0.24\textwidth}%
  \includegraphics[width=\linewidth]{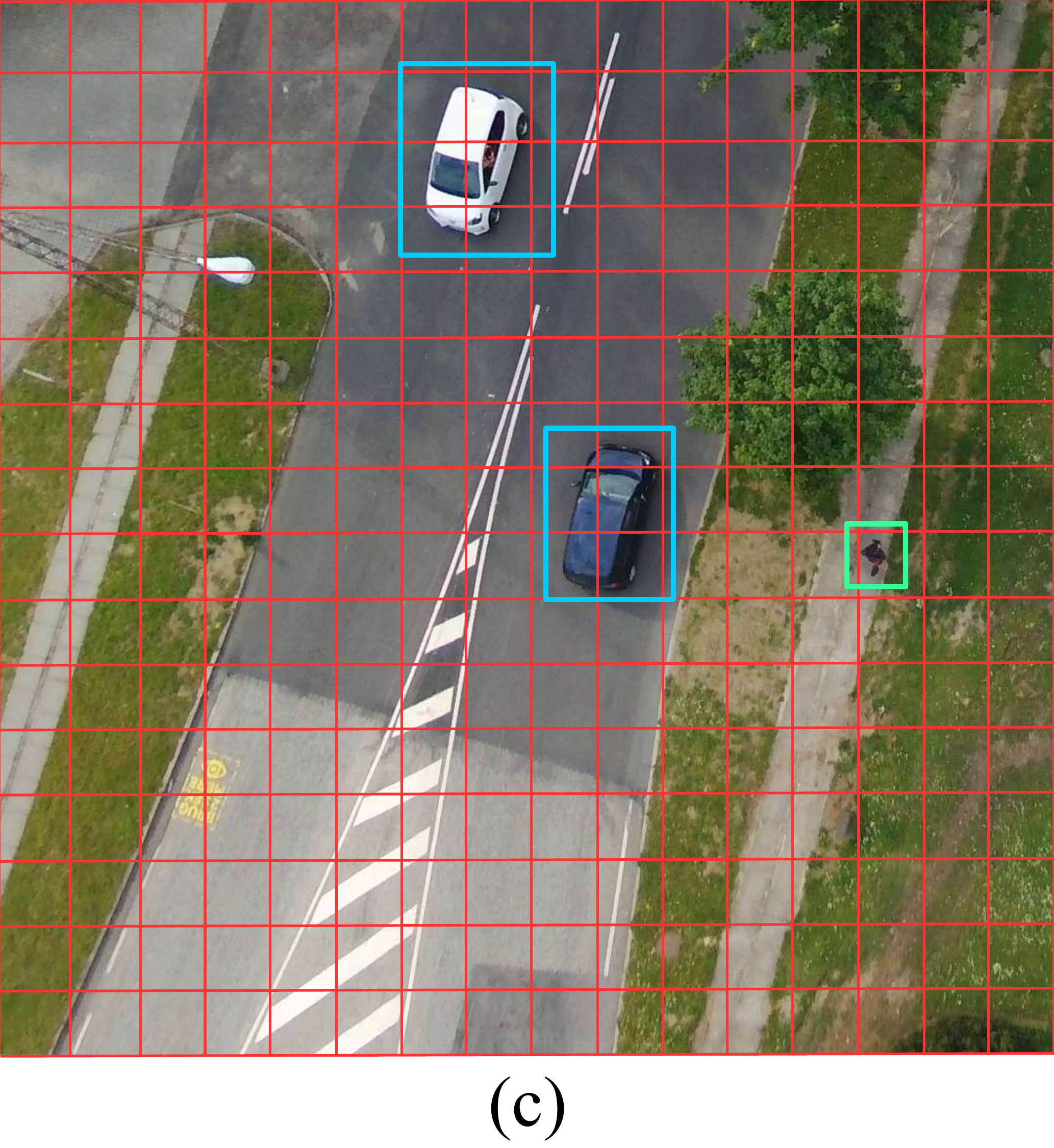}\label{fig:scene_model_c}
\endminipage\hfill
\minipage{0.24\textwidth}%

  \includegraphics[width=\linewidth]{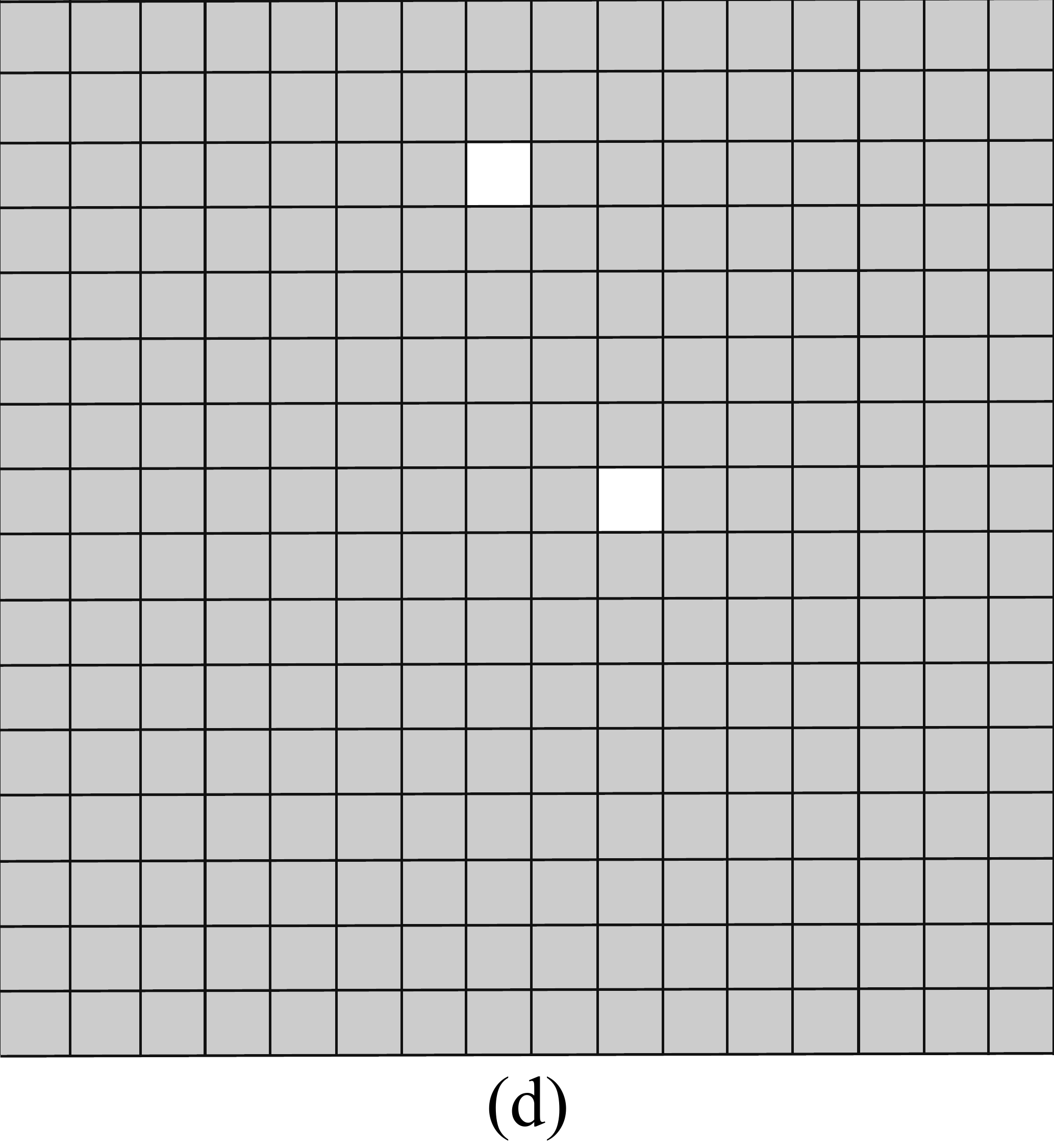}\label{fig:scene_model_d}
\endminipage
\caption{This illustration shows how we get the environment representation from the raw image. Objects in the input image (a) are annotated by bounding boxes using off-the-shelf object detector methods or human supervision (b). Then, the image is split into $16\times16$ grid cells (c) and cells, including object bounding box centers, are set to value 1, otherwise set to 0 (d). Note that the sub-figure (d) is a matrix for the car category, and only shows car instances. The metadata extraction process can be done for all types of objects given in object vocabulary $O$.}\label{fig:scene_model}
\end{figure*}

Let us consider an aerial image dataset with annotated object bounding boxes ($\mathbf{D}$) including $N$ images $\{\mathbf{I}_n\}^{N}_{n=1}$, with $\mathbf{I} \in {R}^{d_x \times d_y \times d_z}$ where $d_x$ and $d_y$ represent the number of pixels in $x$ and $y$ direction and $d_z$ represents the number of color channels. We can divide the image into grids and assign a value to each grid cell according to the presence or absence of a particular object. Therefore, we can use a new metarepresentation ($\mathbf{X}$) of the image dataset ($\mathbf{D}$). A binary tensor $\mathbf{X}_i$ represents objects and their locations explicitly in input image $\mathbf{I}_i$, which has the form:
\begin{equation}
    \mathbf{X} \in \{0,1\}^{N_x \times N_y \times N_o},  
\end{equation}
where $N_x$ and $N_y$ are the numbers of grid cells along $x-$ and $y-$ axis respectively, and $N_o$ is the number of objects in object vocabulary $O$, i.e., the number of object types that the object detector can detect. Each grid cell has the width $s_x$ and the height $s_y$, where $s_x  =  ({d_x}/{N_x})$ and $s_y  =  ({d_y}/{N_y})$. In other words, a grid tensor $\mathbf{X}$ consists of $N_o$ grid matrices and each matrix consists of $N_x$ columns and $N_y$ rows. An element of a matrix has a value of $1$, if the corresponding grid area includes center of a particular object; otherwise it has a value of $0$. The overall process can be seen in Fig. \ref{fig:scene_model}.

In this work, we crop aerial images to a size of $1080\times1080$ pixels and divide them into $16\times16$ grids, in which each grid has a size of $67 \times 67$ pixels. The grid size is chosen according to the average bounding box size of human instances (i.e., the smallest object category in the dataset) so that only one human instance can occupy a grid cell.

\subsection{UAV-AdNet -- Deep Neural Network for Anomaly Detection}

In UAV-AdNet, we use a DNN architecture that learns objects distribution in an environment representation conditioned on GPS label corresponding to the environment. Moreover, we constrain the network to have continuous latent space which allows the network to learn the data distribution. This constraint is crucial in the context of anomaly detection task since the network can reason about anomaly objects when they are out of the distribution.

\subsubsection{Input form}
Firstly, we flatten a binary tensor $\mathbf{X}$ to a binary vector ($\mathbf{x}$) which has a length of $N_x \times N_y \times N_o$. Then, we create a vector $\mathbf{l}$ $\in R^2$ for the GPS label of the environment, including latitude and longitude as scalar elements. We feed both $\mathbf{x}$ and $\mathbf{l}$ as input to the model.

%
%

\subsubsection{Network architecture}

\begin{figure*}[!hbt] 
\centerline{
\includegraphics[width=0.98\textwidth]{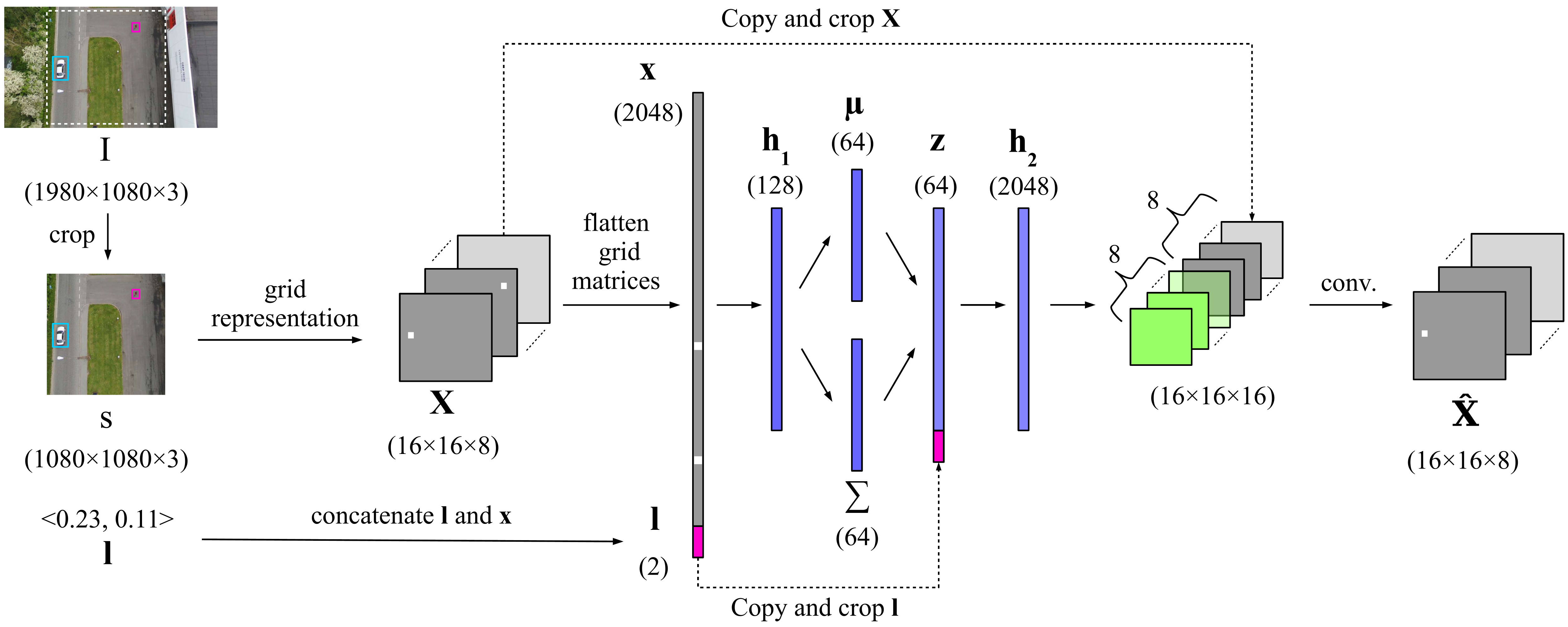}
}
\caption{This illustration shows the proposed network architecture and anomaly detection process. Firstly, RGB images are captured by UAV (I), and objects are annotated using off-the-shelf object detectors. Then, the region of interest is cropped (s), and the heuristic-based method is applied to get the grid representation of the environment ($\mathbf{X}$). The network takes both grid representations ($\mathbf{X}$) and GPS labels ($\mathbf{l}$) as input. Then, it predicts the reconstructed sample ($\mathbf{\hat{X}}$) which cells, include the anomaly objects that differ from the original input.\label{fig:architecture}}
\end{figure*}

As shown in Fig. \ref{fig:architecture}, UAV-AdNet takes a binary grid vector and GPS label as input and outputs a new grid vector ($\mathbf{\hat{x}}$) in which the grid cells including anomalies is set to 0.

\begin{algorithm}[hb!]
\caption{Anomaly detection process with UAV-AdNet.}\label{alg:inference}
\begin{algorithmic}[1]
\State \textbf{Input:} Test image $\mathbf{i}$, GPS label $\mathbf{l}$
\State \textbf{Output:} Detected anomalies.
\State

\State Predict the object detector output $O$ for given input $\mathbf{i}$.
\State Get the grid representation $\mathbf{x}$ from $O$.
\State Run forward-pass of UAV-AdNet with $\mathbf{x}$ and $\mathbf{l}$ to get the reconstructed sample $\mathbf{\hat{x}}$.
\State Compare $\mathbf{x}$ and $\mathbf{\hat{x}}$ to find anomalies.

\end{algorithmic}
\end{algorithm}

The architecture of UAV-AdNet is given Fig \ref{fig:architecture}. At high level, UAV-AdNet can be seen as a sequence of layers that process the input data at different scales. The first dense layer has 128 hidden neurons and is followed by ReLU activation. The second dense layer consists of two independent groups of hidden neurons represented by $\boldsymbol{\mu}$ and  $\boldsymbol{\Sigma}$, which learn mean and variance of the probability distribution of input data, respectively. Then, the data are fed into the sampling layer, which is a deterministic parameterized transformation of a prior distribution \cite{kingma2013auto}. After sampling the latent vector ($\mathbf{z}$), it is concatenated with the GPS input vector ($\mathbf{l}$). The combined vector is fed into the Dense3 layer, which has $2048$ hidden neurons, and the ReLU non-linearity function is applied. The outputs of ReLU3 are reshaped to $16 \times 16 \times 8$ sized tensor and concatenated with tensor $\mathbf{X}$. Consequently, $1 \times 1$ sized convolutional kernels are applied and $\mathbf{\hat{X}}$ is calculated as output. The whole process can be summarized in Alg. \ref{alg:inference}.

During our experiments, we observe that the crop-and-copy link for tensor $\mathbf{X}$ increases the network performance significantly. We interpret this observation as that the copy-and-crop link carries information from the input layer, directly, to the last layer. Therefore, even if the inputs are quite sparse matrices (as in our case), the network can recall them using the copy-and-crop link.


\subsubsection{Loss function}
In order to learn the parameters of a probability distribution, which represents the input data instead of just compressed representation of it, we use VAE-like loss function ($\mathcal{L}$):
\begin{equation}
   \mathcal{L} = \mathcal{L}_{rec} + \mathcal{L}_{reg}, \label{eqn:overall}
\end{equation}
which consists of reconstruction loss ($\mathcal{L}_{rec}$) and a regularizer term ($\mathcal{L}_{reg}$).

We use binary cross-entropy loss (log loss) as reconstruction loss, which penalizes the network for being dissimilar of the input grid matrices ($\mathbf{X}$) and the output grid matrices ($\mathbf{\hat{X}}$). Therefore, reconstruction loss can be defined as:
\begin{equation}
    \mathcal{L}_{rec} = - \frac{1}{N} \sum_{i=1}^N {\mathbf{x_i} \log{f(\mathbf{ x_i}, \mathbf{ l_i}) }}+(\mathbf{1}-\mathbf{x_i})\log{(\mathbf{1}-f(\mathbf{ x_i}, \mathbf{ l_i}))},
\end{equation}
where $N$ is the number of samples in the dataset, $f(\mathbf{x_i}, \mathbf{l_i})$ is the model output for given inputs $\mathbf{x_i}$ and $\mathbf{l_i}$, and $\mathbf{1}$ is the one vector which has the same shape with $x_i$.

We add a constraint on the encoding part as in VAE, which forces it to generate latent vectors that roughly follow a Gaussian unit distribution. We can interpret this phenomenon as the network can learn object distributions in the environment if it is constrained to have continuous latent space. The following regularizer term is added to the overall loss function (\ref{eqn:overall}) to add the constraint:
\begin{equation}
    \mathcal{L}_{reg} = {{D}_\text{KL}}[q(\boldsymbol{\mu}, \boldsymbol{\Sigma} \mid \mathbf{x}, \mathbf{l})\mid\mid p(\mathbf{z})],   
\end{equation}
where $D_\text{KL}$ is the Kullback-Leibler divergence which measures the discrepancy between two probability distribution, $q(\boldsymbol{\mu}, \boldsymbol{\Sigma} \mid \mathbf{x}, \mathbf{l})$ is the probability distribution which is represented by model parameters $\boldsymbol{\mu}$, $\boldsymbol{\Sigma}$ for given $\mathbf{x}$ and $\mathbf{l}$, and $p(\mathbf{z})$ is the prior distribution for latent variables ($\mathbf{z}$).

By assuming latent prior is given by Gaussian distribution as in VAEs, i.e., $p(\mathbf{z})=\mathcal{N}(\boldsymbol{\hat{\mu}}, \boldsymbol{\hat{\Sigma}})$ where $\mathcal{N}$ is the Gaussian distribution with the mean vector $\boldsymbol{\hat{\mu}}=\vec{0}$ and the variance matrix $\boldsymbol{\hat{\Sigma}}=I$, the regularizer term can be re-written as follows:
\begin{align}
\mathcal{L}_{reg}
&= \frac{1}{2}\left[-\log{|\boldsymbol{\Sigma}|} - n_h + \text{tr} \{ \boldsymbol{\Sigma} \} + \boldsymbol{\mu}^T \boldsymbol{\mu}\right]\\
&= \frac{1}{2}\left[-\sum^{n_h}_i\log\sigma_i^2 - n_h + \sum^{n_h}_i\sigma_i^2 + \sum^{n_h}_i\mu^2_i\right],
\end{align}
{\noindent}where $n_h$ is the number of hidden neurons in $\mathbf{\sigma}$ and $\mathbf{\mu}$ latent layers; $\mu_i$ and $\sigma_i$ are mean and variance hidden neurons indexed by $i$; $tr$ is the trace of a square matrix. 

\subsubsection{Training settings}
For training the networks, we use Adam optimizer, whose parameters are empirically set as learning rate is $0.001$, beta1 is $0.9$, beta2 is $0.999$. Training data are split into mini-batches with a batch size of 64. The training process is finished when validation accuracy begins to decrease.

\subsection{Dataset Collection}
There are several aerial image and video datasets collected by UAVs (\cite{zhuvisdrone2018, du2018unmanned}), and they are used for computer vision tasks such as object detection, image segmentation. However, they do not include any data which are related to the environment except for images and object annotations. Therefore, they do not allow different data modalities, which is very relevant to robotics. For this reason, we collected our dataset, which includes only bird-view images and corresponding GPS label and flight data (e.g., IMU, battery level).

\begin{figure} 
\centerline{
\includegraphics[width=0.49\textwidth]{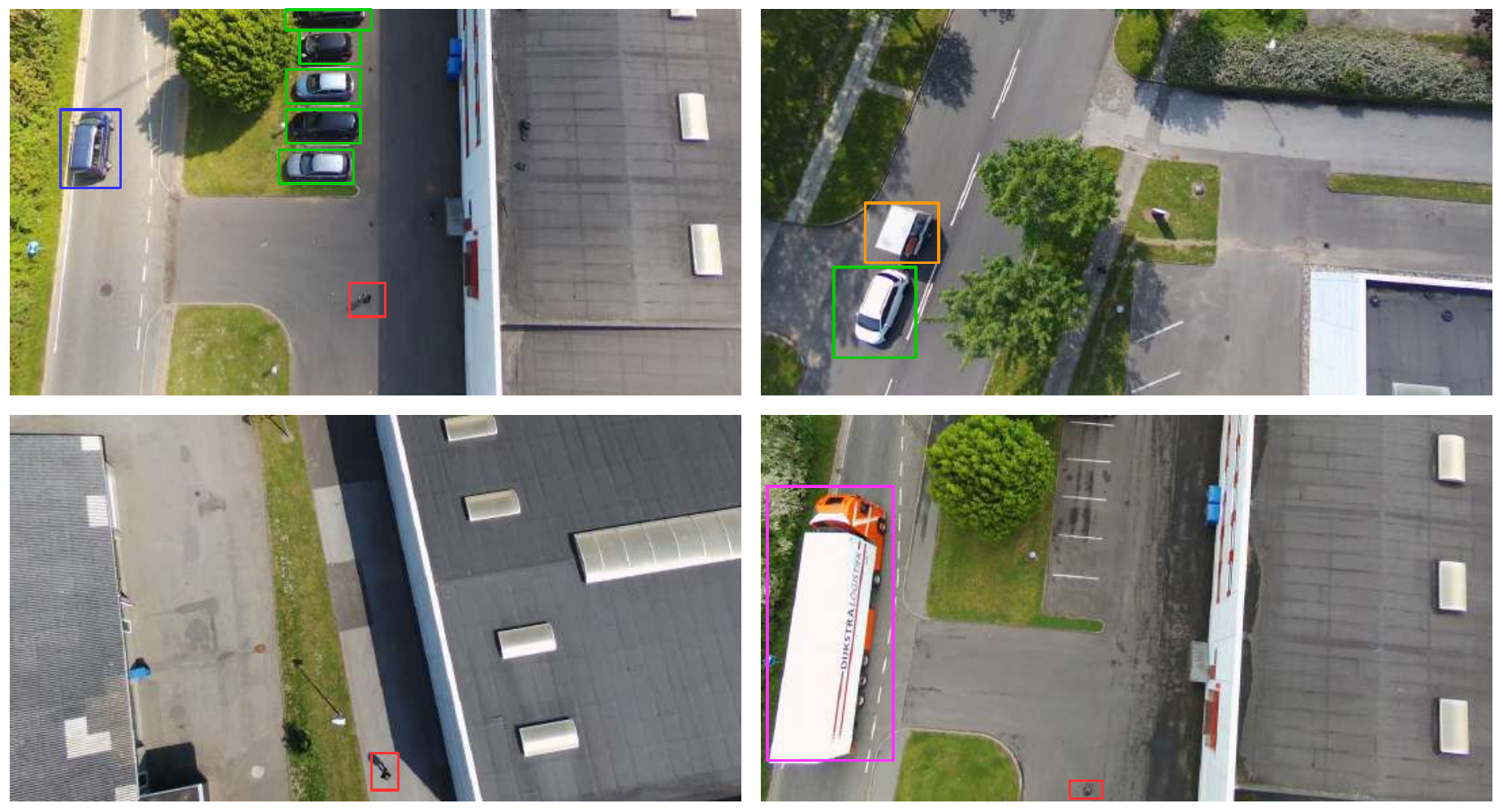}
}
\caption{Samples from the dataset. Objects are annotated with bounding boxes. Scenes may contain multiple instances belonging same class (upper left), composed vehicles such as a car with a trailer (upper right), significantly small objects such as humans, bikes (lower left) and huge objects such as trucks that occupy a large area of a scene (lower right).\label{fig:samples_dataset}}
\end{figure}

For dataset collection, we use Parrot Bebop 2 drone that has a fish-eye camera that allows capturing bird-view images. The dataset is collected around Aarhus University DeepTech Experimental Hub in Aarhus (Denmark) (See Figure \ref{fig:samples_dataset}). The drone followed a predefined path around the building with the 30-meter altitude under different weather conditions (e.g., sunny, partially sunny, cloudy, windy), and recorded video, flight data and GPS data simultaneously. We have $15000$ samples in total and split the whole dataset into three categories: 60\% for training, 10\% for validation and 30\% for testing.

\section{EXPERIMENTS AND RESULTS}
In this section, we evaluate and compare our method with several baselines for detection of different types of anomalies. For this end, we separate test set into three categories according to included anomaly types:
\begin{itemize}
    \item Existence of an object which breaks private rules of the environments (e.g., human in the forbidden zone of the building)
    \item Existence of an object which breaks public rules (e.g., car on the pedestrian road).
    \item Existence of an object which does not break neither public nor private rule yet arises suspicions due to its rare observation (e.g. truck in the car parking area).
\end{itemize}

We compare methods in terms of reconstruction capacity and anomaly detection capability.

\subsection{Baselines}
Although there are classical machine learning algorithms (principal component analysis \cite{wold1987principal}) that are used for anomaly detection, we compare our method with deep learning-based state-of-the-art models. For this end, we choose autoencoders and its variants as a baseline for anomaly detection tasks. 


In order to observe the effect of GPS input on the network performance, we create a variant of UAV-AdNet just by removing the GPS input layer and call it ``UAV-AdNet-wo-gps''. Secondly, we remove copy-and-crop connections for input grids and keep the GPS input layer. Therefore, the network has a form of conditional variational autoencoder (CVAE). Lastly, we use vanilla autoencoder (VAE) by removing the copy-and-crop connections and the GPS input layer. For the sake of compatibility, all models have the same number of hidden layers and neurons as in UAV-AdNet. 

We trained all networks with the same batch size and optimizer parameters until the validation error started to increase.

\subsection{Evaluation Metrics}
For evaluating the reconstruction capability of the methods, we use precision (Pr), recall (Rec) and F1-score (F1), which are defined as follows \cite{bozcan2019cosmo}:
\begin{equation}
\textrm{Pr}  =  \frac{\textrm{TP}}{\textrm{TP}+\textrm{FP}}  \qquad
\textrm{Rec}  =  \frac{\textrm{TP}}{\textrm{TP}+\textrm{FN}} \qquad
\textrm{F1}  =  2\cdot \frac{\textrm{Pr}\cdot \textrm{Rec}}{\textrm{Pr} + \textrm{Rec}} \label{eqn:metrics},
\end{equation}
{\noindent}where TP, FP and FN stand for the number of true positives, false positives and false negatives, respectively. For anomaly detection task, we define TP as the number of grid cells that have value of $1$ in reconstructed data correctly according to the ground truth sample; FP as the number of grid cells that the model set to $1$ but should have been $0$ according to the ground truth; TN as the number of grid cells that are set to $0$ correctly according to ground truth, and FN as the number of grid cells that the model is set to $0$, yet should have been $1$ according to the ground truth.

For evaluating the anomaly detection performance of the methods, we define accuracy as the percentage of anomalies correctly estimated for the labeled anomalies in the test dataset.

\subsection{Comparing Reconstruction Capabilities of Models}
Abnormal objects (i.e., anomalies) are detected comparing original input and the reconstructed data. Besides finding anomalies in input data, in the reconstructed sample, models should be able to recall ordinary (not anomaly) objects in the original input as well. Moreover, the reconstructed data should not contain further objects which differ from the original input. In other words, the models should have high precision and recall.

By the definitions given in (\ref{eqn:metrics}), the reconstruction performances of the models are given in Table \ref{tbl:reconstructionperformance}. UAV-Adnet and UAV-AdNet-wo-gps have significantly better precision and recall values compared to CVAE and VAE, which shows that combining input data and hidden representation (via copy-and-crop connections) of the environment increases the model performance significantly. Moreover, we observe that feeding the model with GPS inputs also increases the model's reconstruction.

The copy-and-crop connections increase the reconstruction performance of the model significantly. They carry information regarding the original input to the last layer. These connections prevents the vanishing of the input information through layers.

\begin{table}[hbt]
\caption{Reconstruction performances of the methods over the test set. \label{tbl:reconstructionperformance}}
\centering
\footnotesize
\begin{tabular}{|l|ccc|}\hline                                                                 
&\textbf{Precision}  & \textbf{Recall}  & \textbf{F1-score}  \\ \hline \hline
UAV-AdNet & \textbf{0.9816}	& \textbf{1.0}   & \textbf{0.9907}   \\ 
UAV-AdNet-wo-gps & 0.9427 & 0.9984  &  0.9697  \\ 
CVAE & 0.1963 & 0.5165  &  0.2845   \\ 
VAE  & 0.1920	 &  0.4102   &  0.2616 \\ \hline
  
\end{tabular}
\end{table}

\subsection{Task 1: Finding Anomaly Objects Breaking Private Rules}

UAV-AdNet can find anomaly objects for the special rules of the DeepTech Experimental Hub building. According to these rules, anomaly cases are defined as follows:
\begin{itemize}
    \item Occurrence of any human or vehicle at the backside of the building.
    \item Occurrence of any vehicle at the left side of the building.
\end{itemize}
For testing, an object, that breaks a private rule, is added to the grid representations (i.e., corresponding grid cells are set to 1) of the test samples in order to create these anomaly cases. Then, modified grid representation and corresponding GPS label are fed into the model, and the model is expected to set grid cells, including anomaly objects to 0 in the reconstructed sample. 

We evaluate this task with VAE, CVAE, UAV-AdNet-wo-gps and our model, as shown in Table \ref{tbl:accuracy}. We see that our model provides the highest accuracy for Task 1. When we remove the GPS input layer from the model (UAV-AdNet-wo-gps), the anomaly detection performance decreases. Moreover, we observe that removing copy-crop connections for grid inputs decreases the accuracy significantly.

\subsection{Task 2: Finding Anomaly Objects Breaking Public Rules}
In this task, we can test our model to find anomalies breaking public rules. Similar to Task 1, firstly, public rules are defined as follows:
\begin{itemize}
    \item Pedestrians can cross a road using zebra crossings only.
    \item Bike can ride on bike road only.
    \item Vehicles are not allowed to ride or park on a bike road except bicycles and motorbikes.
\end{itemize}
According to these rules, anomaly objects are added to grid representations as in Task 1. After that, these objects are founded by comparing original input and reconstructed input during the model's reconstruction phase. As shown in Table \ref{tbl:accuracy}, our model performs better than UAV-AdNet-wo-gps, CVAE and VAE for Task 2. We can see that feeding the network with GPS data increases the accuracy and it performs poorly when copy-crop layers are removed.

\subsection{Task 3: Finding Suspicious Objects}
Lastly, we evaluate the models to find an abnormal object even if it does not break private or public rules. These objects may be still suspicious since they barely occur in an environment representation with a given GPS coordinate. For instance, the occurrence of a person on the roof of the building does not break rules but might arise a suspicion.

For this task, we add random objects to the grid representations according to the context of the scene and without breaking public or private rules. As shown in Table \ref{tbl:accuracy}, our model provides the highest accuracy for Task 3 as well. However, results for this task relatively worse than Task 1 and 2 since the models tend to keep the added object in the reconstructed sample if the added object suits the contextual layout of previous observations (e.g. if a car is added to a road then the model may not consider it as anomaly).

\begin{table}[hbt]
\caption{Anomaly detection accuracy. \label{tbl:accuracy}}
\centering
\footnotesize
\begin{tabular}{|l|ccc|}\hline                                                                 
&\textbf{Task 1}  & \textbf{Task 2}  & \textbf{Task 3}  \\ \hline \hline
UAV-AdNet & \textbf{0.9214}	& \textbf{0.8614}   & \textbf{0.8255}   \\ 
UAV-AdNet-wo-gps & 0.8778 & 0.7912  &  0.7778  \\ 
CVAE & 0.6212 & 0.4413  &  0.4433   \\ 
VAE  & 0.6198	 &  0.4165   &  0.4200 \\ \hline
  
\end{tabular}
\end{table}

\section{DISCUSSION}
We observe that the crop-and-copy link for the input layer increases the network's reconstruction performance significantly. By this architecture, UAV-AdNet recovers the original input (high recall and precision), and remove anomalies from the reconstructed sample. Moreover, we observe that feeding GPS to UAV-AdNet also increases the reconstruction and anomaly detection performances, whereas it is not as effective as crop-and-copy links for inputs.

We also observe that feeding the network with irrelevant GPS labels during the inference phase reduces the reconstruction performance significantly. This phenomenon is expected since UAV-AdNet learns conditional distribution of grid representations for given GPS labels. Therefore, it can find anomalies for the environments which have similar object layouts but different GPS layouts.

Since collecting test set for anomaly detection and labeling objects as an anomaly is not a straightforward task, we add anomaly objects randomly according to specific rules (e.g., public, private, suspicious) which creates anomaly cases. However, random adding of objects can create regular environment layouts which do not contain anomaly cases. Therefore, as future work, a more comprehensive test set can be collected, including annotated anomalies.

In our experiments, the drone flies at a constant altitude (30 meters) to capture data. Therefore, the grid representation of an environment might be different if the flight altitude changes. For instance, when the drone flies at a higher altitude, objects have a smaller appearance. In that case, the grid size should be decreased.

The camera position may not be perpendicular to the Earth when the drone is accelerating or has a non-zero roll and pitch attitude. In that case, a gimbal for the camera can be used to fix the camera position since grid representations should be extracted from complete bird-view images.

To create test cases, we add objects randomly into the environment representations according to their contextual layout. Although these synthetic samples are consistent with real samples, they may cause an unforeseen bias in the dataset. Hence, we will enlarge the dataset with real samples in order to prevent the bias of synthetic samples. 

Although we have conducted our experiments in a warehouse-like building (Aarhus University DeepTech Experimental Hub), the proposed method is applicable for different types of structured environments such as harbors, factories, airports. To this end, after collecting a dataset in a specific zone using predefined flight paths, a UAV can operate for the anomaly detection task. Moreover, it can record new samples (both aerial images and GPS data) during the inspection for further training later on.

In this work, we only consider GPS data to increase anomaly detection performance. Different types of flight data can be used as input to indicate the current context of the environment as future work. Our results indicating the effect of GPS data are promising to use different data modalities to increase anomaly detection performance.

\section{CONCLUSIONS}

In this paper, we have proposed a heuristic method for environment representation of bird-view images and a DNN-based anomaly detection method (UAV-AdNet) trained on environment representations and GPS labels jointly. In our experiments, we show that the proposed architecture has better scene reconstruction performance with the copy-and-crop connection for the input. Moreover, compared to autoencoder variants, which are heavily used for anomaly detection in different domains, we show that our model performs better on several anomaly detection tasks. We also observe that feeding the network with GPS data can augment anomaly detection performance. 






\bibliographystyle{IEEEtran}
\bibliography{references}

\end{document}